\def\year{2021}\relax
\documentclass[letterpaper]{article} 
\usepackage{aaai21}  
\usepackage{times}  
\usepackage{helvet} 
\usepackage{courier}  
\usepackage[hyphens]{url}  
\usepackage{graphicx} 
\usepackage{natbib}  
\usepackage{caption} 
\nocopyright
\usepackage[utf8]{inputenc} 
\usepackage[T1]{fontenc}    
\usepackage{hyperref}       
\usepackage{url}            
\usepackage{booktabs}       
\usepackage{amsfonts}       
\usepackage{nicefrac}       
\usepackage{microtype}      
\usepackage{amsmath}
\usepackage{amssymb}
\usepackage{amsthm}
\usepackage{mathtools}
\usepackage{Definitions}
\usepackage{natbib}
\usepackage[normalem]{ulem}
\usepackage[linesnumbered, ruled, vlined]{algorithm2e}
\usepackage{todonotes}
\usepackage{enumitem}
\usepackage{multirow}
\usepackage[switch]{lineno}

\newtheorem{defin}{Definition}

\title{Robust Fairness under Covariate Shift}
\author{Ashkan Rezaei\textsuperscript{\rm 1},
Anqi Liu\textsuperscript{\rm 2},
Omid Memarrast\textsuperscript{\rm 1},
Brian Ziebart\textsuperscript{\rm 1} \\
}

\affiliations{
    \textsuperscript{\rm 1} University of Illinois at Chicago \\
    \textsuperscript{\rm 2} California Institute of Technology\\ 
    arezae4@uic.edu, anqiliu@caltech.edu,
    omemar2@uic.edu, bziebart@uic.edu 
}

\begin{document}
\maketitle

\begin{abstract}
Making predictions that are fair with regard to protected attributes (race, gender, age, etc.) has become an important requirement for classification algorithms. Existing techniques derive a fair model from sampled labeled data relying on the assumption that training 
and testing 
data are identically and independently drawn (iid) from the same distribution.
In practice, distribution shift can and does occur between training and testing datasets as the characteristics of individuals interacting with the machine learning system change.
We investigate fairness under covariate shift, a relaxation of the iid assumption in which the inputs or covariates change while 
the conditional label distribution remains the same. 
We seek fair decisions under these assumptions on target data with unknown labels. 
We propose an approach that obtains the predictor that is robust to the worst-case 
testing performance while satisfying target fairness requirements and matching statistical properties of the source data. 
We demonstrate the benefits of our approach on benchmark prediction tasks.
\end{abstract}

\section{Introduction}

Supervised learning algorithms typically focus on optimizing one singular objective: predictive performance on unseen data.
However, 
the social impact of unwanted bias in these algorithms has become increasingly important. 
Machine learning systems that disadvantage specific groups are less likely to be accepted and may violate disparate impact law \cite{chang2006applying,kabakchieva2013predicting,lohr2013big,shipp2002diffuse,obermeyer2016predicting,moses2014using,shaw1988using,carter1987assessing,o2016weapons}.
Fairness through unawareness, which simply denies knowledge of protected group membership to the predictor, is insufficient to effectively guarantee fairness because other characteristics or covariates may correlate with protected group membership \cite{pedreshi2008discrimination}. 
Thus, there has been a surge of interest in the machine learning community to define fairness requirements reflecting desired behavior and to construct learning algorithms that more effectively seek to satisfy those requirements in various settings \cite{mehrabi2019survey,barocas2017fairness,Calmon2017,donini2018empirical,dwork2012fairness,dwork2017decoupled,hardt2016equality,zafar2017fairness,Zemel13,jabbari2016fair,chierichetti2017fair}. 

Though many definitions and measures of (un)fairness have been proposed (See \citet{verma2018fairness,mehrabi2019survey}),
the most widely adopted are group fairness measures of \emph{demographic parity} 
\cite{calders2009building}, 
\emph{equalized opportunity}, and \emph{equalized odds} \cite{hardt2016equality}. 
Techniques have been developed as either post-processing steps \cite{hardt2016equality} or in-processing learning methods \cite{agarwal2018reductions,zafar2017fairness,rezaei2020fairness} seeking to achieve fairness according to these group fairness definitions. These methods attempt to make fair predictions at testing time by 
strongly assuming that training and testing data are \emph{independently and identically drawn (iid)} from the same distribution, so that providing fairness on the training dataset  provides approximate fairness on the testing dataset.  

In practice, it is common for 
data distributions to \emph{shift} between the training data set (\emph{source distribution}) and the testing data set (\emph{target distribution}). 
For example, the characteristics of loan applicants may differ significantly over time due to macroeconomic trends or changes in the self-selection criteria that potential applicants employ.
Fairness methods that ignore such shifts may satisfy definitions for fairness on training samples, while violating those definitions severely on testing data.
Indeed, disparate performance for underrepresented groups in computer vision tasks has been attributed to manually labeled data that is highly biased compared to testing data
\cite{yang2020towards}. 
Explicitly incorporating these shifts into the design of predictors is crucial for realizing fairer applications of machine learning in practice.
However, the resulting problem setting is particularly challenging; 
access to labels is only available for the training distribution.
Fair prediction methods could fail by using only source labels, especially for fairness definitions that condition on ground-truth labels, like equal opportunity. 

\begin{figure}[!ht]
\begin{tabular}{r l}
\setlength{\tabcolsep}{0pt}
\includegraphics[width=.51\columnwidth, trim ={0cm, 0cm, 0cm, 0cm}, clip]{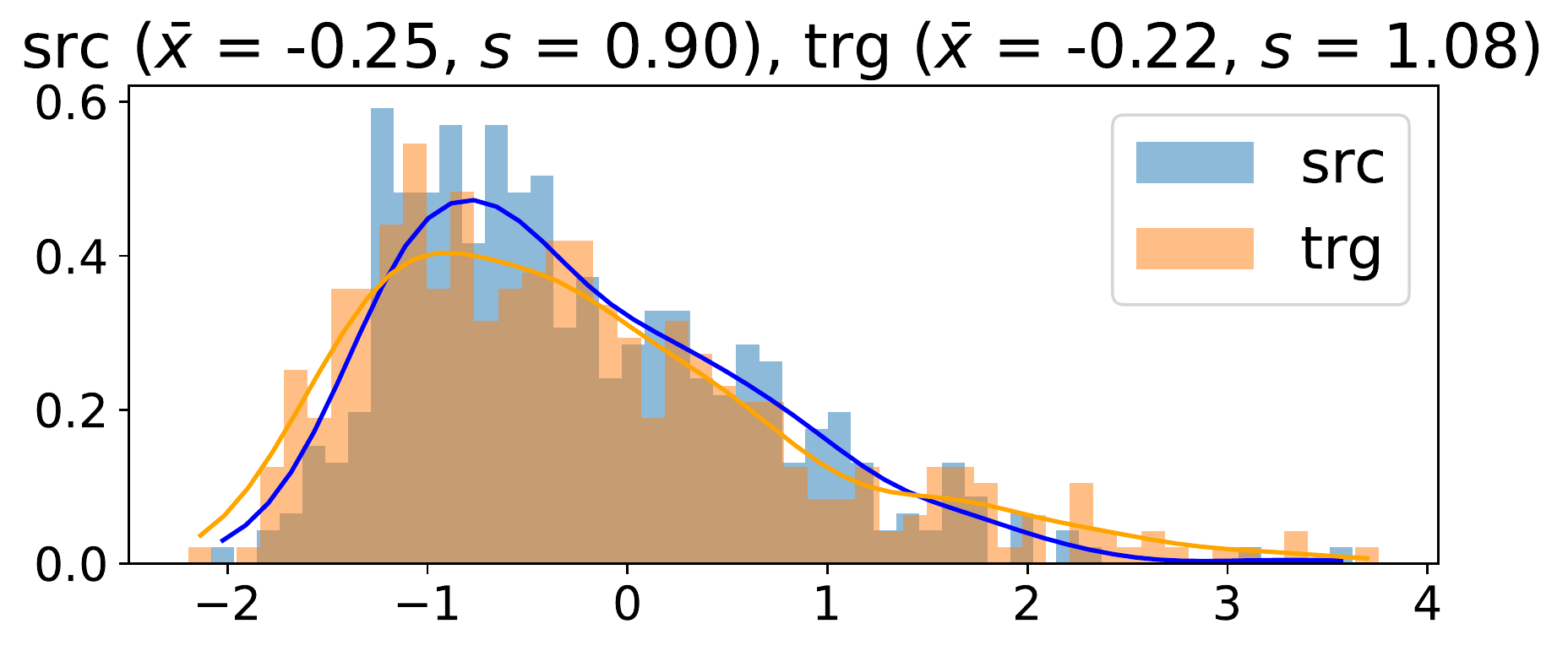} & 
\multirow{3}{*}[1.35cm]{\includegraphics[width=.47\columnwidth, trim ={1.8cm, 0cm, 0cm, 0cm}, clip]{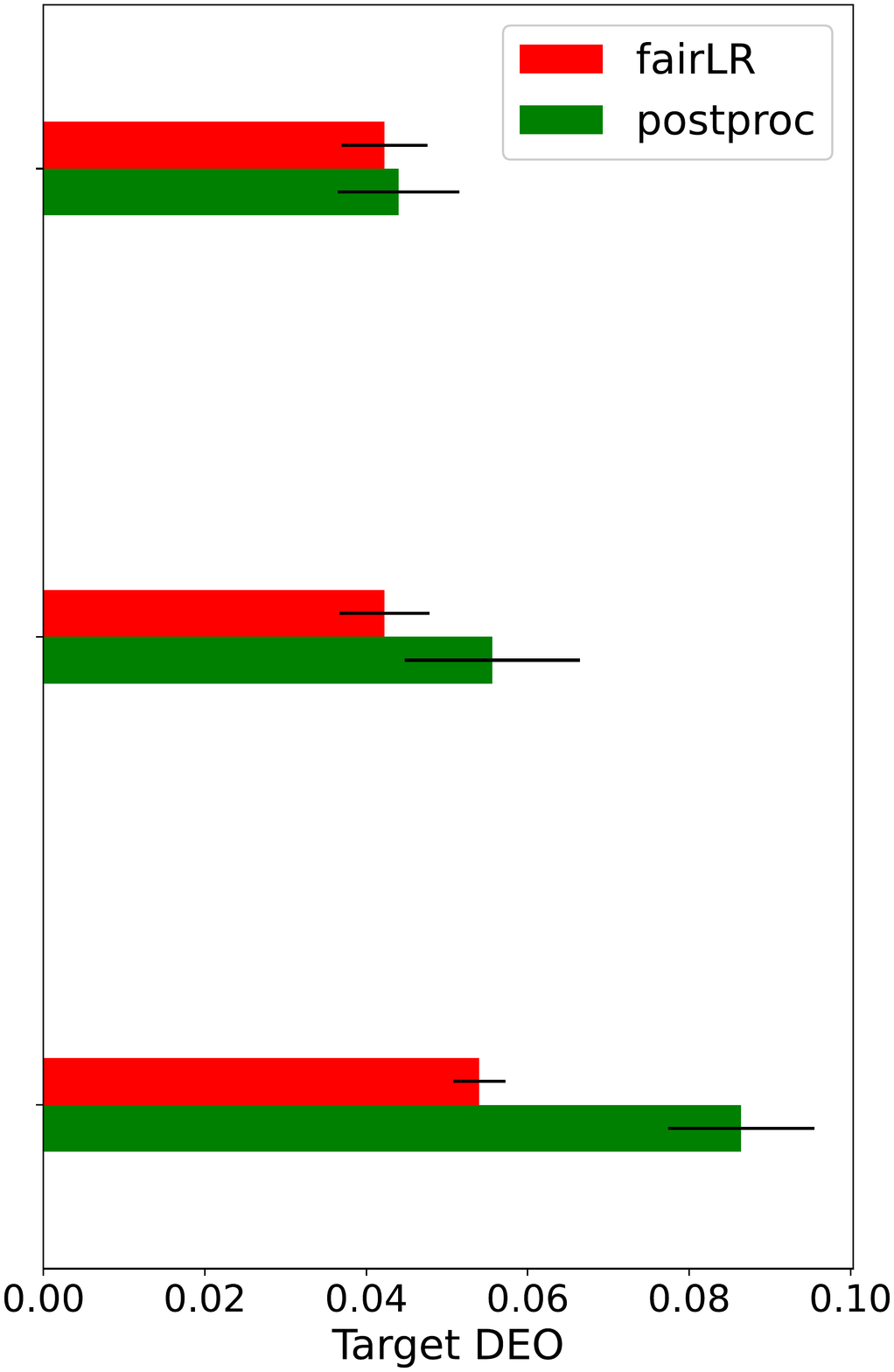}} \\
\includegraphics[width=.5\columnwidth, trim ={0cm, 0cm, 0cm, 0cm}, clip]{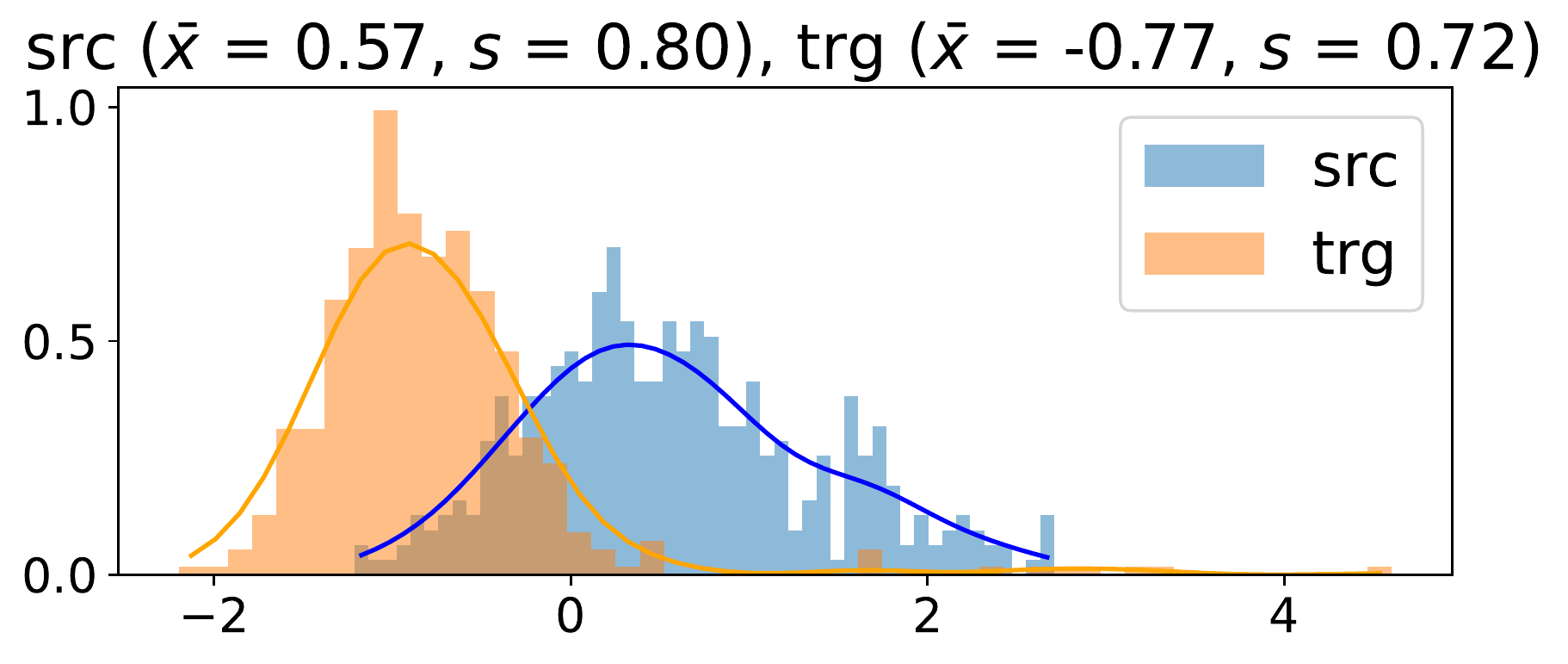} & \\
\includegraphics[width =.5\columnwidth ]{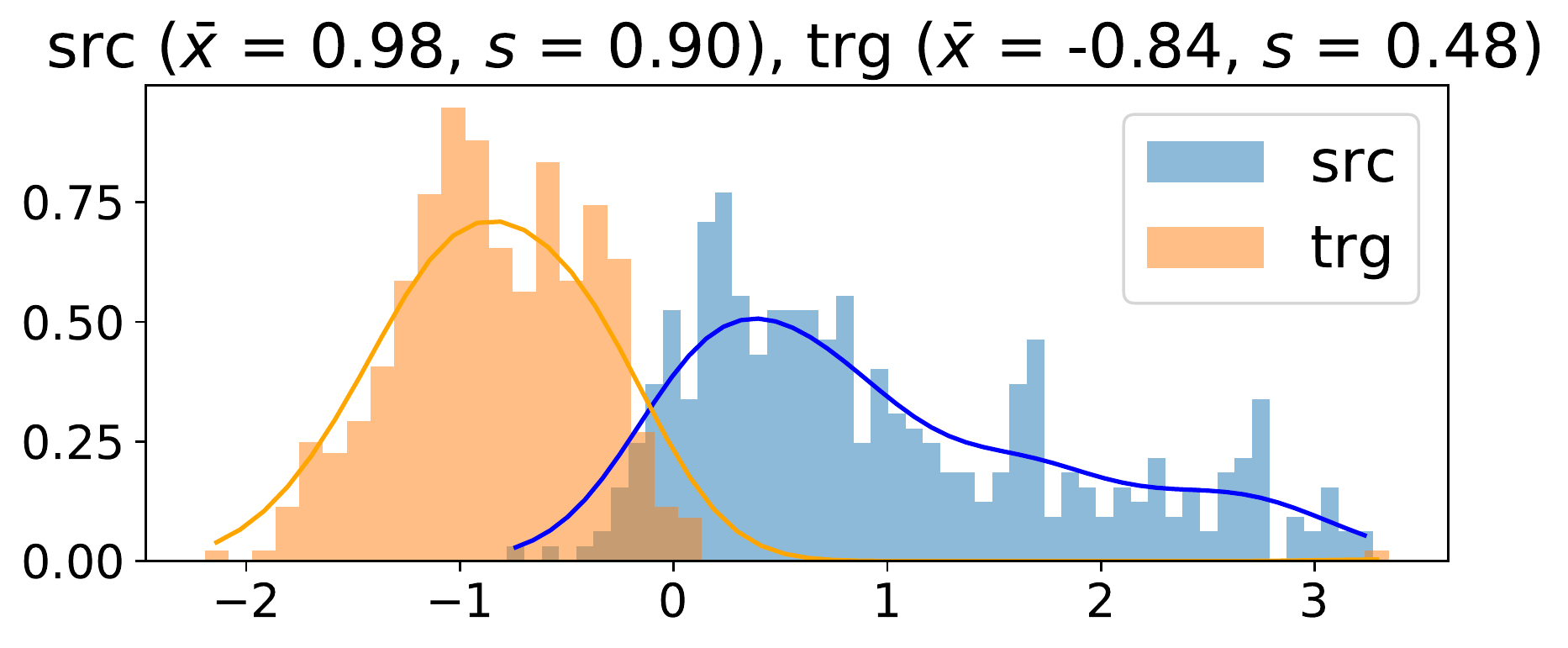} & 
\end{tabular}
\caption{The difference of equalized opportunity between two genders in German UCI dataset evaluated on the target distribution for (a) the post-processing method of \citet{hardt2016equality} and (b) an in-processing fairLR method \cite{rezaei2020fairness}, that both do not account for distribution shift, 
and correct true positive rate parity on source data. The histograms on the left show the corresponding distribution shift on first principal component of the covariates between source and target data.
The shift intensity has an overall increasing effect on DEO violation of both methods. Experiment section includes details on sampling procedure.}
\label{fig:motivation}
\end{figure}

Figure \ref{fig:motivation} illustrates the declining performance of a post-processing method  \cite{hardt2016equality} and an in-processing method \cite{rezaei2020fairness}  
that do not consider distribution shift and instead only depend on source fairness measurements. 
Therefore, relying on the iid assumption, which is often violated in practice, introduces significant limitations for realizing desired fairness 
in critical applications.

We seek to address the task of providing fairness guarantees under the non-iid assumption of covariate shift. 
Covariate shift is a special case of data distribution shift. It assumes that the relationship between labels and covariates (inputs) is the same for both distributions, while only the source and target covariate distributions differ. Under the fair prediction setting, the sensitive group features are usually correlated with other features. Covariate shift then indicates that the labels given the covariates, including the sensitive group features, stays the same between two distributions. For example, even though there are fewer female loan applicants in area A than area B, which causes a marginal input distribution to shift between these two areas, we believe the probability of belonging to the advantages class (e.g.,  repaying a loan) given the full covariate should be the same. 

In this paper, we propose a robust estimation approach for constructing a fair predictor under covariate shift. We summarize our contribution as follows:
We formulate the fair prediction problem as a game between an adversary choosing conditional label distributions to fairly minimize predictive loss on the target distribution, and an adversary choosing conditional label distributions that maximize that same objective. Constraints on the adversary require it to match statistics under the source distribution. Fairness is incorporated into the formulation by a penalty term in the objective that evaluates the fairness on the target input distribution and adversary's conditional label distribution.
We derive a convex optimization problem from the formulation and obtain the predictor's and adversary's conditional label distribution that are parametric, but cannot be solved analytically. Based on the formulation, we propose a batch gradient descent algorithm for learning the parameters. 
We compare our proposed method with baselines that only account for covariate shift or fairness or both. We demonstrate that our method 
outperforms the baselines on both predictive performance and the fairness constraints satisfaction under covariate shift settings.


%


\section{Related Work}

\paragraph{Fairness}

Various methods have been developed in recent years to achieve fair classification according to group fairness definitions. These techniques can be broadly categorized as pre-processing modifications to the input data or fair representation learning \cite{Kamiran2012, Calmon2017,Zemel13,feldman2015certifying,del2018obtaining,donini2018empirical, guidotti2018survey}, post-processing correction of classifiers' outputs \cite{hardt2016equality,Pleiss17}, in-processing methods that incorporate the fairness constraints into the training process and parameter learning procedure \cite{donini2018empirical,zafar2017aistats,zafar2017fairness,zafar2017parity,cotter2018,goel2018non,woodworth2017learning,kamishima2011fairness,bechavod2017penalizing,quadrianto2017recycling, rezaei2020fairness}, meta-algorithms \cite{celis2019classification,menon2018cost}, reduction-based methods \cite{agarwal2018reductions,cotter2018}, or
generative-adversarial training \cite{madras2018learning, zhang2018mitigating, celis2019improved,xu2018fairgan,adel2019one}.

 
The closest to our work in this category is the fair robust log-loss 
predictor of \citet{rezaei2020fairness}, which operates under an iid assumption. That formulation 
similarly builds a minimax game between a predictor and worst-case approximator of the true distribution. The main difference is that under the iid assumption, the true/false positive rates of target groups can be expressed as a linear constraint on the source data, in which the ground truth label is known. However, in our work, no sample-based constraint is available for measuring true/false positive rates on target data, because the target true label is unknown under the covariate shift assumption. Thus, we enforce these fairness measures by an expected penalty of the worst-case approximation of the target data.

\paragraph{Covariate Shift}

General distribution and domain shift works focus on the joint distribution shift between the training and testing datasets \cite{daume2006domain,ben2007analysis,blitzer2008learning}. Particular assumptions like covariate shift \cite{shimodaira2000improving,sugiyama2007covariate,gretton2009covariate} and label shift \cite{scholkopf2012causal,lipton2018detecting,azizzadenesheli2019regularized} help quantify the distribution shift using importance weights since they introduce invariance in conditional distributions between the training and testing. Importance weighting methods under covariate shift suffer from high variance and sensitivity to weight estimation methods. It has been shown to often be brittle---providing no finite sample generalization guarantees---even for seemingly benign amounts of shift \cite{cortes2010learning}. Applying importance weighting to fair prediction has not been broadly investigated and may suffer from a similar issue. 

Fairness under perturbation of the attribute has been studied by \citet{awasthi2020equalized}. \citet{lamy2019noise} study fair classification when the attribute is subjected to noise according to a mutually contaminated model \cite{scott2013classification}. Our method works for a general shift on the joint distribution of attribute and features, and does not rely on a particular noise model.

Causal analysis has also been proposed for addressing fairness under dataset shift  \cite{singh2019fair}. It
requires a known causal graph of the data generating process, with a context variable causing the shift, as well as the known separating set of features. Our model makes no assumptions about the underlying structure of the covariates. We assume covariate shift, which relates with causal models when there is no unobserved confounders between covariates and labels. 
Given a known separating set of features in the causal model under data shift, the covariate shift assumption holds if we use only the separating set of features for prediction.
Our model builds on robust classification method of \cite{liu2014robust} under covariate shift, where the target distribution is estimated by a worst-case adversary that maximizes the log-loss while matching the feature statistics under source distribution. Therefore, if we know the separating set of features, we can incorporate them as constraints for the adversary. However, it is usually difficult to know the exact causal model of the data generating process in practice.

\section{Approach}
\subsection{Preliminaries \& Notation}


We assume a binary classification task $Y,\Yhat \in \{0,1\}$, where $Y$ denotes the true label, and $\Yhat$ denotes the prediction for a given instance with features $\Xvec \in \Xcal$ and group attribute $A \in \{0,1\}$. We consider $y = 1$ as the privileged class (e.g., an applicant who would repay a loan). 
Further, we assume a given source distribution $(\Xvec, A, Y) \sim P_\src$ over features, attribute, and label, and a target distribution $(\Xvec,A) \sim P_\trg$ over features and attribute only, throughout our paper. 

\subsubsection{Fairness Definitions}
Our model seeks to satisfy the group fairness notions of 
equalized opportunity and odds \cite{hardt2016equality}. 
Our focus in this paper is equalized opportunity, 
which requires equal true positive rates 
 across groups, i.e., for a general probabilistic classifier $P$: 
\begin{align}
P(\Yhat\! =\! 1 | A\! =\! 1, Y\! =\! 1) &= P(\Yhat\! =\! 1| A\! =\! 0, Y\! =\! 1). \label{eq:equalized_opp} 
\end{align}
Our model can be generalized for equal odds, which in addition to providing equal true positive rates across groups, also requires equal false positive rates across groups: 
\begin{align}
P(\Yhat\! =\! 1 | A\! =\! 1, Y\! =\! 0) &= P(\Yhat\! =\! 1| A\! =\! 0, Y\! =\! 0). \label{eq:equalized_odd} 
\end{align}

For Demographic parity \cite{calders2009building} 
which requires equal positive rates across protected groups, i.e., $P(\Yhat = 1 | A = 1) = P(\Yhat = 1| A = 0)$, our model reduces to a special case, as we later explain. 

\subsubsection{Covariate Shift}
In the context of fair prediction, the covariate shift assumption is that the distribution of covariates and group membership can shift between source and target distributions:
\begin{align}
    P_\src(\xvec,a,y) = P_{\src}(\xvec,a) P(y|\xvec,a)\\
    P_\trg(\xvec,a,y) = P_{\trg}(\xvec,a) P(y|\xvec,a).
\end{align}
Note that we do not assume how the sensitive group membership $a$ is correlated with other features $\xvec$. If causal structure between the features and labels were known, as assumed in \citet{singh2019fair}, we could also incorporate a hidden or latent covariate shift assumption. For example, given that there is no unobserved confounder between the covariate and the labels, if $h=\Phi(\xvec, a)$ represents the separating set of features, we can assume $P_{\src}(y|h=\Phi(\xvec, a)) = P_{\trg}(y|h=\Phi(\xvec, a))$ and use $\Phi(\xvec, a)$ instead of $(\xvec, a)$ in our formulation. In this paper, we still use $(\xvec, a)$ to represent the covariates for simplicity. 
\subsubsection{Importance Weighting}
A standard approach for addressing covariate shift is to reweight the source data to represent the target distribution \cite{sugiyama2007covariate}. 
A desired statistic $f(x,a,y)$ of the target distribution can be obtained using samples from the source distribution $(x_i,a_i,y_i)_{i=1:n}$:
\begin{align}
    &\mathbb{E}_{\substack{P_\trg(\xvec,a) \\ P(y|\xvec,a)}}\!\!\left[f(\Xvec,A,Y)\right]
    \approx \sum_{i=1}^n \frac{{P_\trg}(\xvec_i,a_i)}{P_\src(\xvec_i,a_i)}f(\xvec_i,a_i,y_i).\label{eq:importance-weighting}
\end{align}
As long as the source distribution has support for the entire target distribution (i.e., 
$P_{\trg}(\xvec,a) > 0 \implies P_{\src}(\xvec,a) > 0$),
this approximation is exact asymptotically as $n \rightarrow \infty$.
However, the approximation is only guaranteed to have bounded error for finite $n$ if the source distribution's support for target distribution samples is lower bounded \cite{cortes2010learning}: 
     $\mathbb{E}_{P_{\trg}(\xvec, a)}\left[{P_{\trg}(\Xvec, A)}/{P_{\src}(\Xvec, A)} \right] < \infty$.
Unfortunately, this requirement is fairly strict and will not be satisfied even under common and seemingly benign amounts of shift. For example, if source and target samples are drawn from Gaussian distributions with equal (co-)variance, but slightly different means, it is not satisfied.

\subsubsection{Robust Log Loss Classification under Covaraite Shift}
We base our method on the robust approach of \citet{liu2014robust} for covariate shift, which addresses this fragility of reweighting methods.
In this formulation, the probabilistic predictor $\Pbb$ minimizes the log loss on a worst-case approximation of the target distribution provided by an adversary $\Qbb$ that maximizes the log loss while matching the feature statistics of the source distribution: 
\begin{align}
    & \min_{\Pbb(y|\xvec) \in \Delta} \; \max_{\Qbb(y|\xvec) \in \Delta \cap \Xi} \Ebb_{P_\trg(\xvec)\Qbb(y|\xvec)}[-\log \Pbb(Y|\Xvec)] \notag \\
    = & \max_{\Pbb(y|\xvec) \in \Delta \cap \Xi} H_{P_\trg(\xvec)\Pbb(y|\xvec)}(Y|\Xvec),
    \label{eq:robustLogLossCovariateShift}
\end{align}
where a moment-matching constraint set $\Xi = \{\Qbb \,| \, \Ebb_{P_\src(\xvec)\Qbb(y|\xvec)}[\phi(\Xvec,Y)]= \Ebb_{P_\src(\xvec,y)} [\phi(\Xvec,Y)]\}$  on source data is enforced with $\phi(\xvec,y)$ denoting the feature function, and $\Delta$ denoting the conditional probability simplex. A first-order moments feature function, $\phi(\mathbf{x},y) = [x_1y, x_2y, \dots x_my]^\top$, is typical but higher-order moments, e.g.,  $yx_1,yx_2^2,yx_3^n,\dots$ or mixed moments, e.g., $yx_1,yx_1x_2,yx_1^2x_2x_3, \dots$, can be included.
The saddle point solution under these assumptions is $\Pbb = \Qbb$ which reduces the formulation to maximizing the target distribution conditional entropy ($H$) while matching feature statistics of the source distribution. The probabilistic predictor of \cite{liu2014robust} reduces to the following parametric form:
\begin{align}
    \Pbb_\theta(y|\xvec) = {e^{\frac{P_\src(\xvec)}{P_\trg(\xvec)}\theta^\top\phi(\xvec,y)}}\Big/{\sum_{y' \in \Ycal} e^{\frac{P_\src(\xvec)}{P_\trg(\xvec)}\theta^\top\phi(\xvec,y')}}, \label{eq:robust_covariate_paramteric_form}
\end{align}
where the Lagrange multipliers $\theta$ maximize the target distribution log likelihood in the dual optimization problem.

\subsubsection{Robust Log Loss for Fair Classification (IID)} The same robust log loss approach has been employed by \citet{rezaei2020fairness} for fair classification under the iid assumption ($P_\trg = P_\src$), where both the optimization objective and the constraints are evaluated on the source data. Since the true label is available during training, fairness can be enforced as a set of linear constraints on predictor $\Pbb$, which yields a parametric dual form. 

In contrast, under the non-iid assumption, the desired fairness on target cannot be directly inferred by enforcing constraints on the source. Additionally, the true/false positive rate parity constraints are no longer linear because the ground truth label on target data is unobserved. Thus, we replace the target label by a random variable distributed according to the worst-case approximation $\Qbb$, and seek fairness on target by augmenting the objective in \eqref{eq:robustLogLossCovariateShift} with an expected fairness penalty incurred by the worst-case target approximator $\Qbb$. In our formulation, the saddle point solution is no longer simple (i.e., $\Pbb \neq \Qbb$), and no parametric form solution is available.



\subsection{Formulation}

Our formulation seeks a robust and fair predictor under the covariate shift assumption by playing a minimax game 
between a minimizing predictor and a worst-case approximator of the target distribution that matches the feature statistics from the source and marginals of the groups from target. We assume the availability of a set of labeled examples $\{\xvec_i,a_i,y_i\}^n_{i=1}$ sampled from the source $P_\src(\xvec,a,y)$ and unlabeled examples $\{\xvec_i,a_i\}^m_{i=1}$ sampled from target distribution  $P_\trg(\xvec,a)$ during training.  

\begin{defin}
The {\bf Fair Robust Log-Loss Predictor under Covariate Shift}, $\Pbb$ minimizes the worst-case expected log loss with an $\mu$-weighted expected fairness penalty on target, approximated by adversary $\Qbb$ constrained to match source distribution statistics (denoted by set $\Xi$) and group marginals on target ($\Gamma$): 
\begin{align}
    \min_{\Pbb \in \Delta} \; \max_{\Qbb \in \Delta \cap \Xi \cap \Gamma} & \Ebb_{P_\trg(\xvec,a)\Qbb(y|\xvec,a)}[-\log \Pbb(Y|\Xvec,A)] \label{eq:fairRobustLogLossFormulation} \\
    + & \mu \, \Ebb_{P_\trg(\xvec,a)\Qbb(y'|\xvec,a)\Pbb(y|\xvec,a)}[f(A,Y',Y)] \notag
    \end{align}
    such that:
    \begin{align}
    & \Xi(\Qbb):  
    \Ebb_{\substack{P_\src(\xvec,a)\\\Qbb(y|\xvec,a)}}[\phi(\Xvec,Y)] = \Ebb_{P_\src(\xvec,a,y)}[\phi(\Xvec,Y)] \text{ and}\notag 
    \end{align}
    \begin{align}
    & \Gamma(\Qbb):
    \Ebb_{\substack{\small P_\trg(\xvec,a)\\\Qbb(y|\xvec,a)}}[g_k(A,Y)] =  
    \underbrace{\Ebb_{\substack{P_\trg(\xvec,a)\\\Ptil_\trg(y|\xvec,a)}}[g_k(A, Y)]}_{\gtil_k},  \notag
\end{align}
$\forall k \in \{0,1\}$, where $\phi$ is the feature function, $\mu$ is the fairness penalty weight, $g_k(.,.)$ is a selector function for group $k$ according to the fairness definition, i.e., for equalized opportunity: $g_k(A,Y)=\Ibb(A\!=\!k \wedge Y\!=\!1)$ , $\gtil_k$ the estimated group density on target, and $f(.,.,.)$ is a weighting function of the mean score difference between the two groups: 
\begin{align}
    f(A,Y,\Yhat) = 
    \begin{cases}
        \frac{1}{\gtil_1} & \text{if } 
        g_1(A,Y) \wedge \Ibb(\Yhat\!=\! 1) \\
        -\frac{1}{\gtil_0} & \text{if } 
        g_0(A,Y) \wedge \Ibb(\Yhat\!=\! 1) \\
        0   & \text{otherwise.}
    \end{cases}  \label{eq:f}  
\end{align}

\end{defin}
The $\Gamma$ constraint enforces $\Qbb$ to be consistent with the marginal probability of the groups on target ($\gtil_k$) for equalized opportunity (and odds). 
This marginal probability is unknown, since the true label $Y$ on target is unavailable. Thus, we estimate these marginal probabilities by employing the robust model \eqref{eq:robust_covariate_paramteric_form} as $\Ptil_\trg(y|\xvec,a)$ in $\Gamma$ in \eqref{eq:fairRobustLogLossFormulation} to first guess the labels under covariate shift ignoring fairness ($\mu = 0$). We penalize the expected difference in true (or false) positive rate of groups in target according to our worst-case approximation of each example being positively labeled. This needs to be measured on the entire target example set and requires batch gradient updates to enforce.   

Our formulation is flexible for all three mentioned  definitions of group fairness.  For equalized odds, 
a second penalty term for false positive rates ($g_k(A,Y)=\Ibb(A=k \wedge Y=0)$) is required and the corresponding marginal matching constraint in $\Gamma$ needs to be added. For demographic parity ($g_k(A,Y)=\Ibb(A=k)$), because the group definition is independent of the true label, the target groups are fully known and the fairness penalty reduces to a linear constraint of $\Pbb$ on target. In this special case, there is no need for the $\Gamma$ constraint and $\mu$ can be treated as a dual variable for the linear fairness constraint. This reduces to the truncated logistic classifier of \cite{rezaei2020fairness} with the exception that the fairness constraint is formed on the target data. 


We obtain the following solution for the predictor $\Pbb$ by leveraging strong minimax duality \cite{topsoe1979information,grunwald2004game} and strong Lagrangian duality \cite{boyd2004convex}.
\begin{theorem}
\label{thm:fairrobustcovariatepredictor}
Given binary class labels and protected attributes $y,a \in \{0,1\}$, the fair probabilistic classifier for equalized opportunity robust under covariate shift 
as defined in \eqref{eq:fairRobustLogLossFormulation} 
can be obtained by solving: 
\begin{align}
&\log \frac{1 - \Pbb(y = 1|\xvec,a)}{\Pbb(y = 1|\xvec,a)}
+ \mu\,\Ebb_{\Pbb(y'|\xvec,a)}\left[f(a,y = 1,Y')\right] \notag \\
& \quad +\frac{P_{\src}(\xvec,a)}{P_{\trg}(\xvec,a)}\theta^\mathrm{T} \left(\phi(\xvec,y = 1) - \phi(\xvec, y = 0)\right) \label{eq:thm1} \\
& \quad + \!\sum_{k \in \{0,1\}} \lambda_{k} g_{k}(a,y = 1) = 0, \notag
\end{align}
where $\theta$ and $\lambda$ are the dual Lagrange multipliers for source feature matching constraints ($\Xi$) and target group marginal matching ($\Gamma$) respectively, and $\mu$ is the penalty weight chosen to minimize the expected fairness violation on target. 

Given the solution $\Pbb^*$ obtained above, for $\Qbb$ to be in equilibrium (given $\theta$ and $\lambda$) it suffices to choose $\Qbb$ for $y=1$ such that:
$\Qbb(y|\xvec,a) =$
\begin{align}
 \frac{\Pbb^*(y|\xvec,a)}{1 - \mu f(a,y,y) \Pbb^*(y| \xvec,a) + \mu f(a,y,y)\Pbb^{*^2}(y|\xvec,a)}, \label{eq:q-eqopp}
\end{align}
where additionally it must hold that $0 \leq \Qbb(y|x,a) \leq 1 :$
\begin{align*}
    \implies
    \begin{cases}
    0 < \Pbb(y=1|\xvec,a) \leq \frac{1}{\mu f(a,1,1)} & \text{if } \mu f(a,1,1) > 1 \\
    0 < \Pbb(y=1|\xvec,a) < 1 & \text{ otherwise.}
    \end{cases}
\end{align*}
\end{theorem}
Due to monotonicity, $\Pbb$ in \eqref{eq:thm1} is efficiently found using a binary-search in the simplex. 
For proofs and further details, we refer the interested reader to the appendix.




\paragraph{Enforcing fairness}

Our model penalizes the expected fairness violation on the target approximated by worst-case adversary ($\Qbb$). We seek optimal penalty weight $\mu$ by finding the zero-point of expected fairness violation, i.e., $\Ebb_{P_\trg(\xvec,a)\Qbb(y'|\xvec,a)\Pbb(y|\xvec,a)}[f(A,Y',Y)]$ on target (see \eqref{eq:thm1}). 
Under the mild assumption that $\Qbb$ is sufficiently constrained by source feature constraints and target group marginal constraints, the approximated fairness violation by $\Qbb$ is monotone in the proximity of the zero point. 
Thus, we can find the exact zero point by a binary-search in a neighborhood around the zero point.

\paragraph{Learning}

For a given fairness penalty weight $\mu$, our model seeks to learn the dual parameters $\theta$ and $\lambda$, such that the worst-case target approximator ($\Qbb_{\theta,\mu}$) matches the sample feature statistic from the source distribution and the marginal of groups on the target set. Given $\theta^*,\lambda^*$ the solution of \eqref{eq:thm1} obtains the optimal fair predictor $\Pbb^*_{\theta^*,\mu^*}$ which is robust against the covariate shift. 

We employ L2 regularization on parameters $\theta$ 
to improve our model's generalization. This corresponds to relaxing our feature matching constraints by a convex norm. 
We employ a batch gradient-only optimization to learn our model parameters. We perform a joint gradient optimization that updates the gradient of $\theta$ (which requires true label) from the source data and $\lambda$ (which does not require true label) from the target batch at each step. Note that we find the solution to the dual objective of \eqref{eq:fairRobustLogLossFormulation} 
by gradient-only optimization without requiring the explicit calculation of the objective on the target dataset. Hence we only use the density ratios when calculating \eqref{eq:thm1}. The gradient optimization converges to the global optimum because the dual objective is convex in $\theta$ and $\lambda$.    

Given an optimal $\Qbb^*$ from \eqref{eq:q-eqopp}, a set of labeled source samples $\Ptil_\src(\xvec,a,y)$ and unlabeled target samples $\Ptil_\trg(\xvec,a)$, the gradient of our model parameters is calculated as follows:
{\small\begin{align}
    &\grad_\theta \Lcal^\trg(\Pbb,\Qbb,\theta,\lambda) = \Ebb_{\substack{\Ptil_\src(\xvec,y)\\\Qbb(\yhat|\xvec,a)}}\phi(\Xvec,\Yhat) - \Ebb_{\Ptil_\src(\xvec,a,y)}\phi(\Xvec,Y)  \notag
    \\
    &\grad_{\lambda_{a'}} \Lcal^\trg(\Pbb,\Qbb,\theta,\lambda) = \Ebb_{\substack{\Ptil_\trg(\xvec,a)\\\Qbb(y|\xvec,a)}}[g_{k}(A,Y)] - \gtil_k.
\end{align}}%
Note that although the gradient of $\theta$ can be updated stochastically, the gradient update for $\lambda$ relies on calculating the $\Qbb$ marginal for each group on the target batch.  
This process is described in detail in Algorithm \ref{algo:opt3loss}.
\begin{algorithm}
    \newcommand{\LineComment}[1]{\Statex \hfill\textit{#1}}
    \SetKwInput{KwInput}{Input}
    \SetKwInput{KwOutput}{Output}
    \KwInput{Datasets $D_\src = \{\xvec_i,a_i,y_i\}^n_{i=1}$, $D_\trg = \{\xvec_i,a_i\}^m_{i=1}$, ratios $\frac{P_\src}{P_\trg}$, feature function $\phi(\xvec,y)$, decaying learning rate $\eta_t$ }
    \KwOutput{$\theta^*,\lambda^*$}
    $\theta \gets$ random initialization \; 
    $\lambda_a \gets 0$ \;
    \Repeat{convergence}{
    Compute $\Pbb,\Qbb$ for all source dataset examples by finding solution to \eqref{eq:thm1} and \eqref{eq:q-eqopp}
    
    $\nabla_\theta \Lcal \gets \frac1n\sum_{i=1}^n\phi(\xvec_i,y_i) - \frac1n\sum_{i=1}^n\sum_{y \in \Ycal}\Qbb(y|\xvec_i,a)\phi(\xvec_i,y)$ \; 
    
    $\nabla_{\lambda_{k}} \Lcal \gets \gtil_{k} - \frac1m\sum_{i=1}^m\sum_{y \in \Ycal}\Qbb(y|\xvec,a)g_{k}(a,y)$ \; 
    
    $\theta \gets \eta_t(\nabla_\theta\Lcal + C\nabla\|\theta\|)$ 
    
    $\lambda \gets \eta_t(\nabla_\lambda\Lcal)$ \;
    }
    \caption{Batch gradient update for Fair Robust Log-Loss learning under Covariate Shift}
    \label{algo:opt3loss}
\end{algorithm}

\section{Experiments}
\label{sec:exps}
We demonstrate the effectiveness of our method on biased samplings from four benchmark datasets:

\begin{itemize}
    \item The {\tt COMPAS} criminal recidivism risk assessment dataset \cite{larson2016we}. The task is to predict recidivism of a defendant based on criminal history.
    \item UCI {\tt German} dataset \cite{Uci2017}. The task it to classify good and bad credit according to personal information and credit history.
    \item UCI {\tt Drug} dataset \cite{fehrman2017five}. The task is to classify type of drug consumer by personality and demongraphics.
    \item UCI {\tt Arrhythmia} dataset \cite{Uci2017}. The task is to distinguish between the presence and absence of cardiac arrhythmia. 
\end{itemize}
Table \ref{tab:datasets} summarizes the statistics of each dataset.


\paragraph{Biased Sampling:} We model a general shift in the distribution of covariates between source and target, i.e., $P_\src(\xvec, a) \neq P_\trg(\xvec, a)$, by creating biased sampling based on the principal components of the covariates. 
We follow the previous literature on covariate shift \cite{gretton2009covariate} and take the following steps to create the covariate shift on each dataset: We normalized all non-categorical features by z-score. We retrieve the first principal component $\Ccal$ of covariates $(\xvec,a)$ by applying principal component analysis (PCA). We then estimate the mean $\mu(\Ccal)$ and standard deviation $\sigma(\Ccal)$, and set a Gaussian distribution $D_t(\mu(\Ccal),\sigma(\Ccal))$ for random sampling of target. We choose parameters $\alpha,\beta$ and set another Gaussian distribution 
$D_s(\mu(\Ccal) + \alpha, \frac{\sigma(\Ccal)}{\beta})$ for random sampling of source data. We fix the sample size for both source and target to $40\%$ of the original dataset; and 
construct the source data by sampling without replacement in proportion to $D_s$, 
and the target data by sampling without replacement from the remaining data in proportion to $D_t$.

\begin{table}[]
\centering
\begin{tabular}{lrrl}
\hline
Dataset    & \multicolumn{1}{c}{$n$} & \multicolumn{1}{c}{Features} & \multicolumn{1}{c}{Attribute} \\ \hline
{\tt COMPAS}     & 6,167                   & 10       & Race      \\
{\tt German}     & 1,000                   & 20       & Gender    \\
{\tt Drug}       & 1,885                   & 11       & Race      \\
{\tt Arrhythmia} & 452                     & 279      & Gender    \\ \hline
\end{tabular}
\caption{Dataset characteristics.}
\label{tab:datasets}
\end{table}

\begin{figure*}[!ht]
\begin{center}
    \setlength{\tabcolsep}{0pt}
    \begin{tabular}{c c c}
    \includegraphics[width=.3\textwidth]{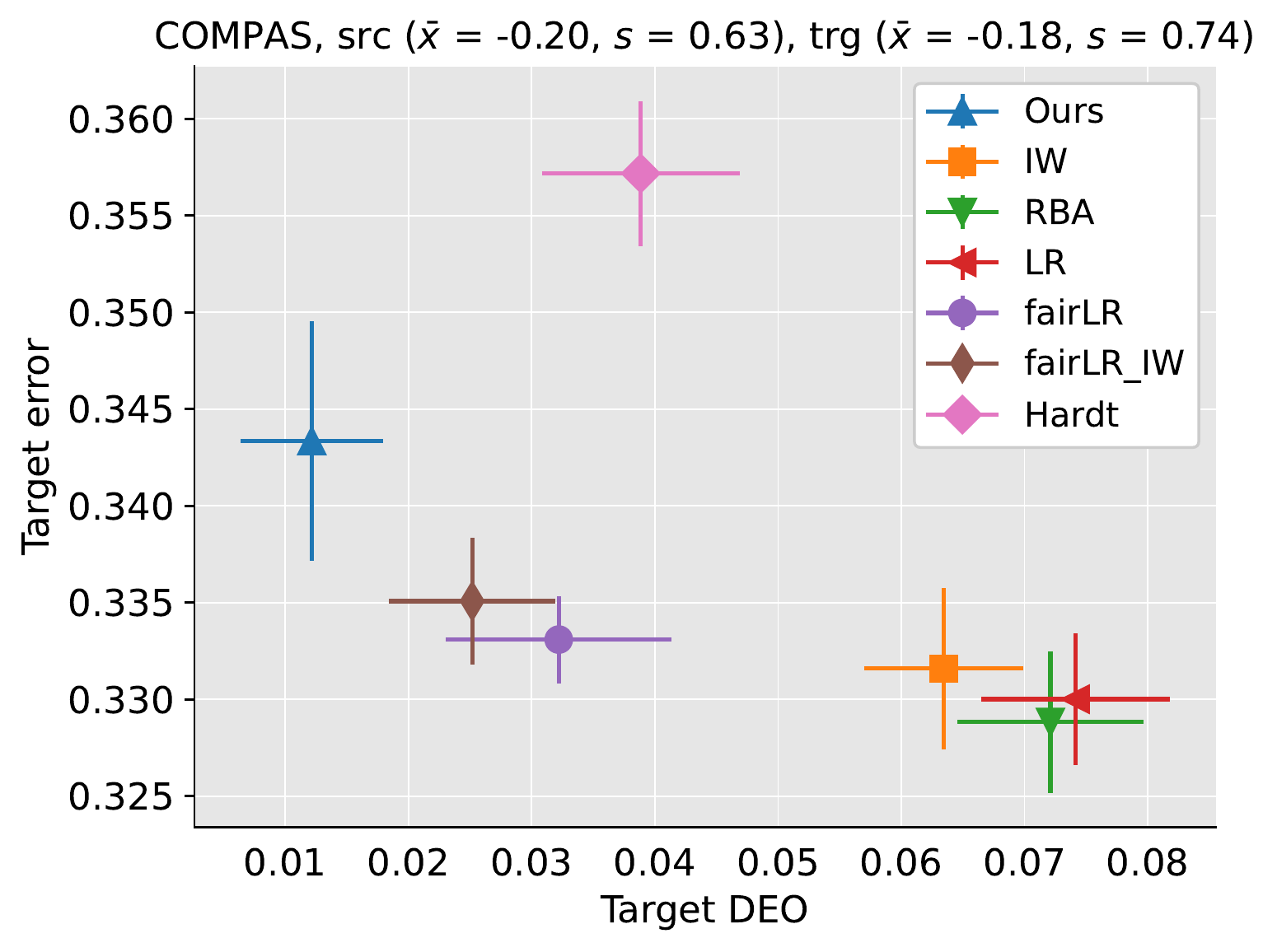} &
    \includegraphics[width=.28\textwidth,trim=20 0 0 0,clip]{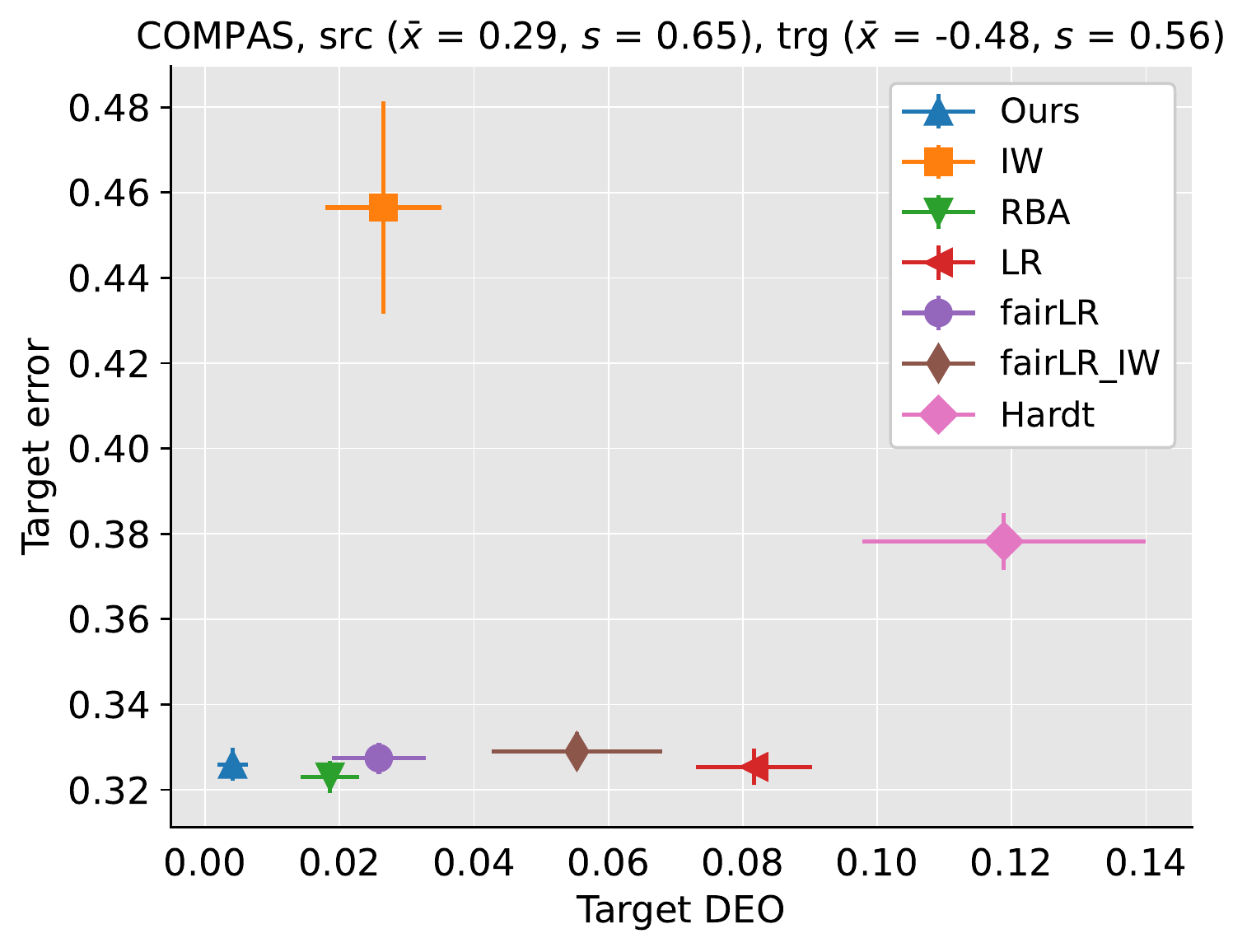} &
    \includegraphics[width=.28\textwidth,trim=20 0 0 0,clip]{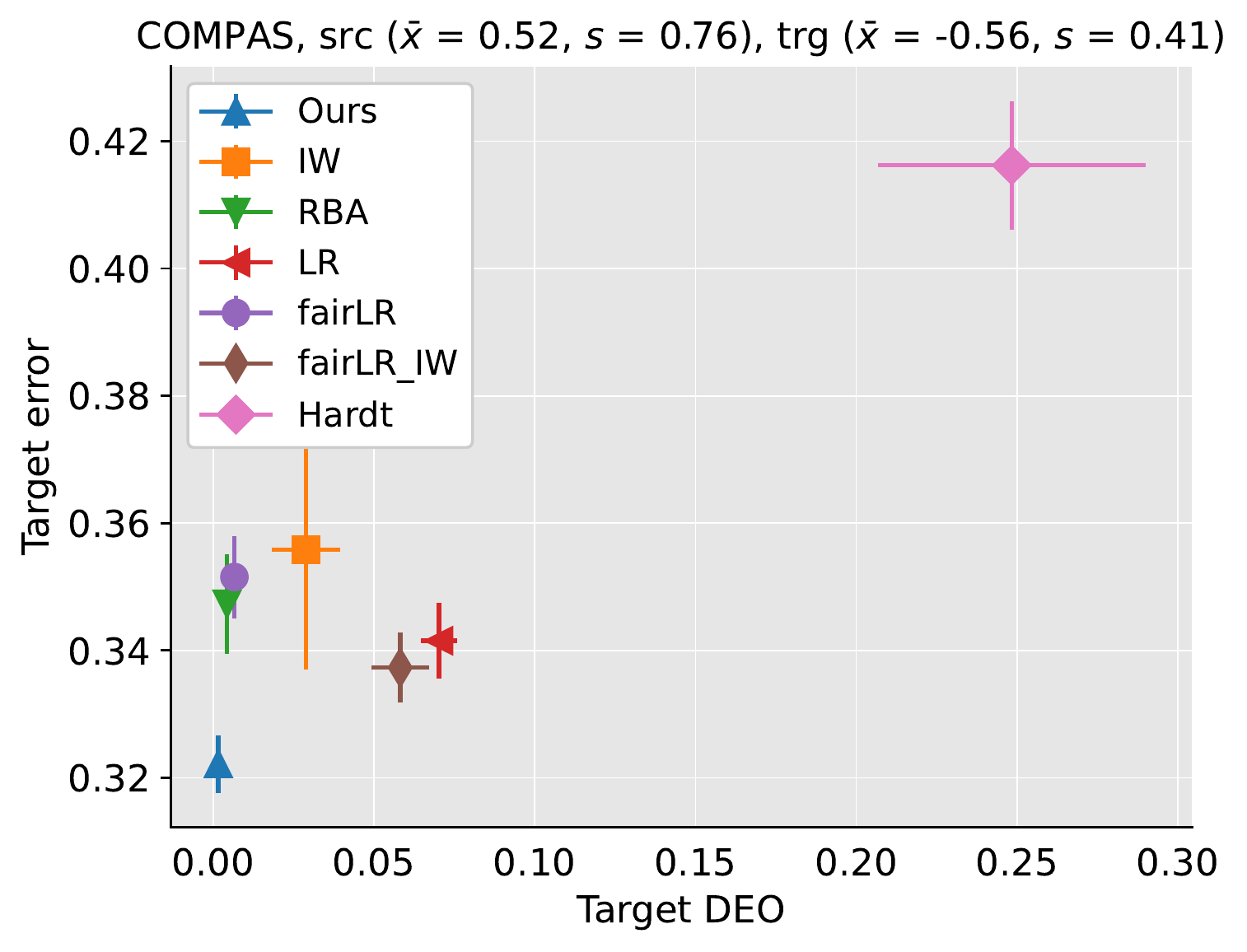}
    \\
    \includegraphics[width=.3\textwidth]{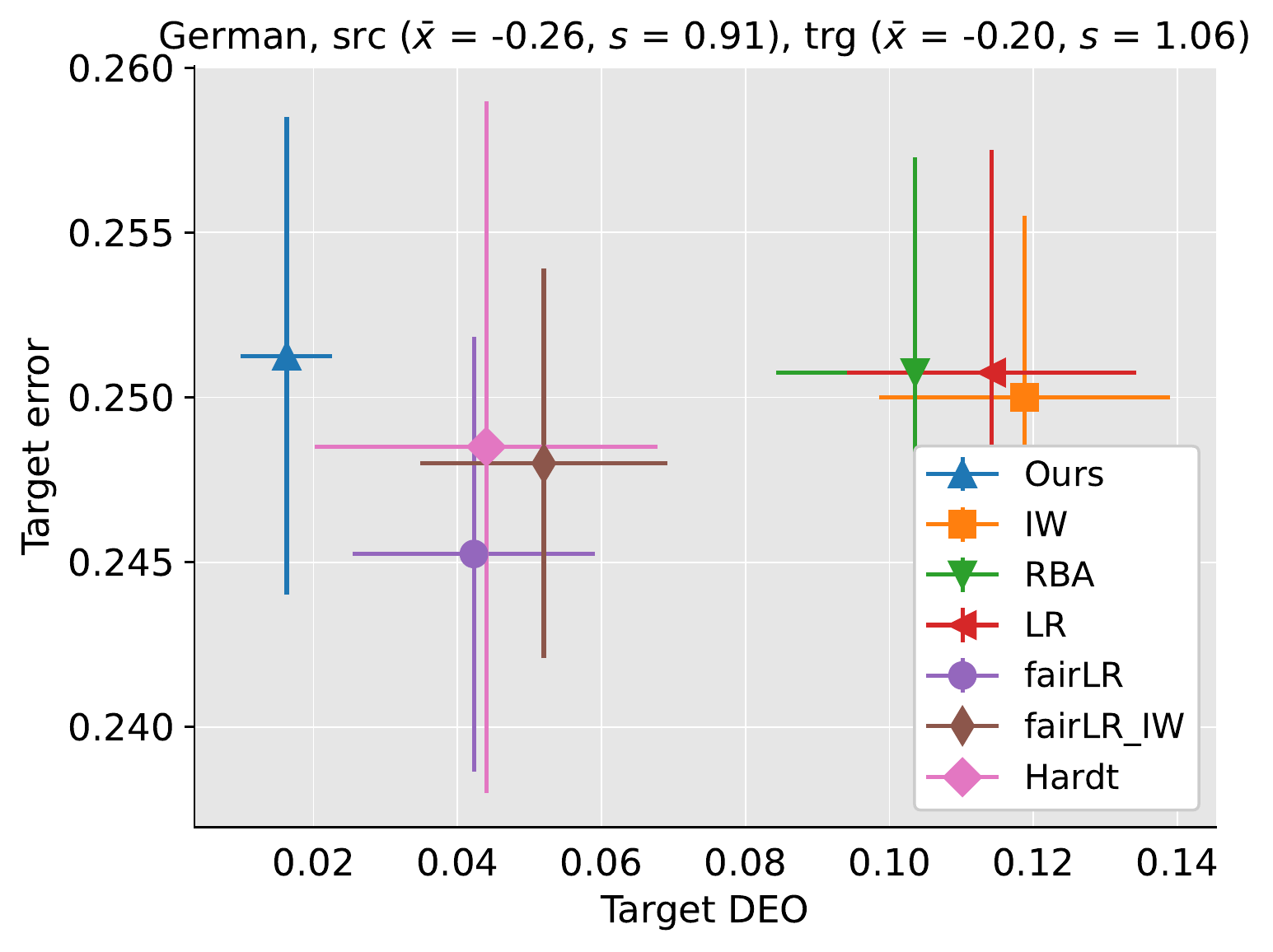} &
    \includegraphics[width=.28\textwidth,trim=20 0 0 0,clip]{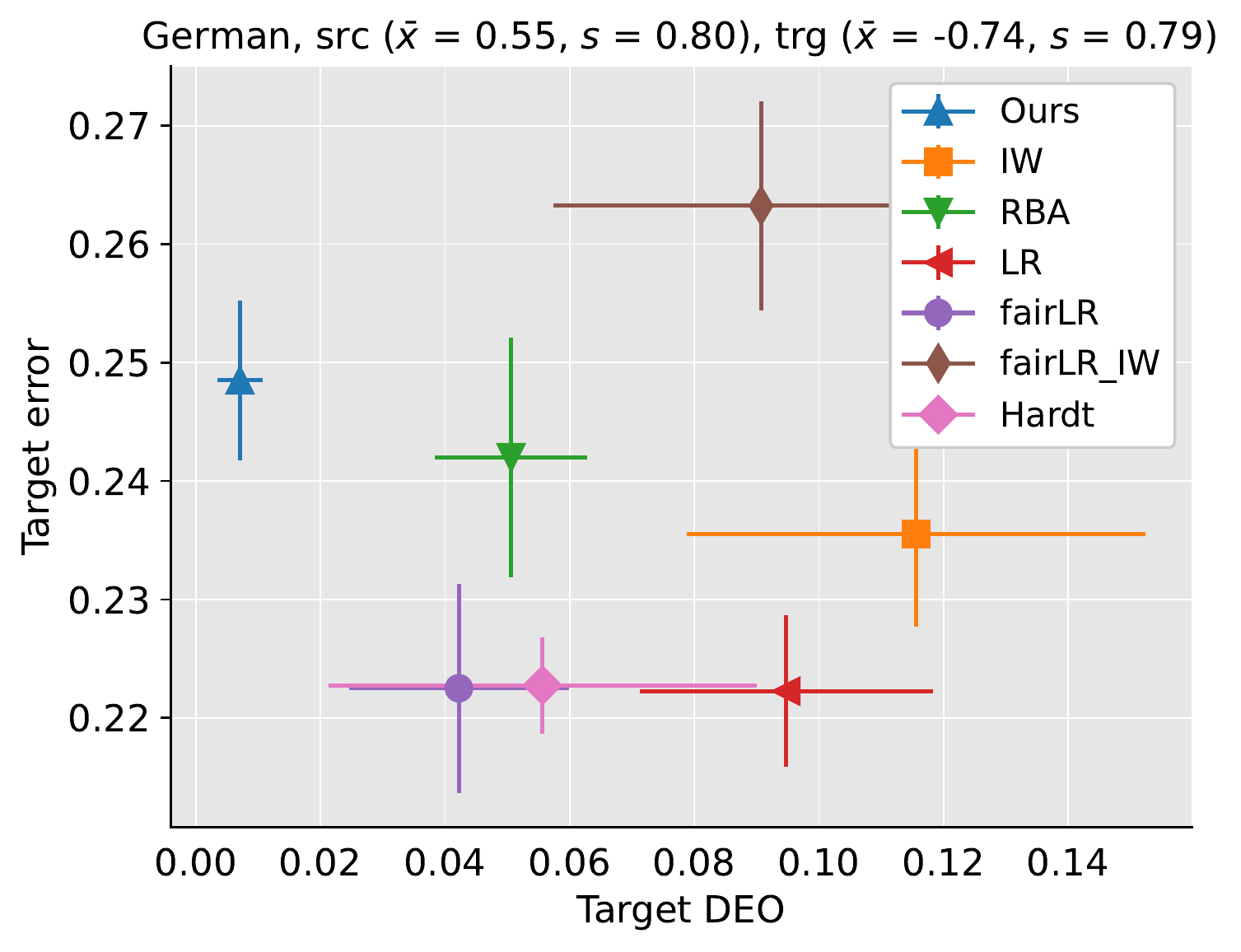} &
    \includegraphics[width=.28\textwidth,trim=20 0 0 0,clip]{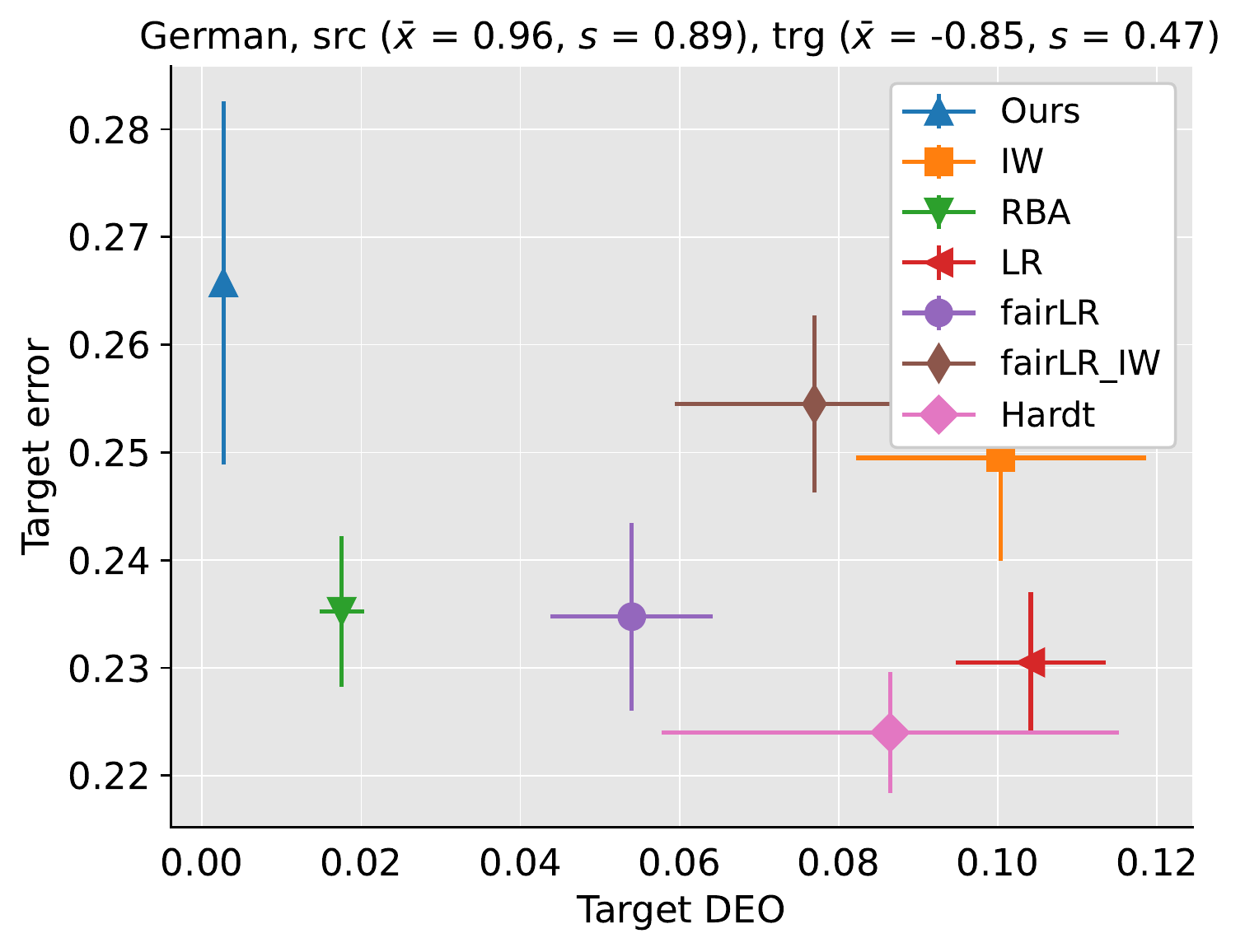}
    \\
    \includegraphics[width=.3\textwidth]{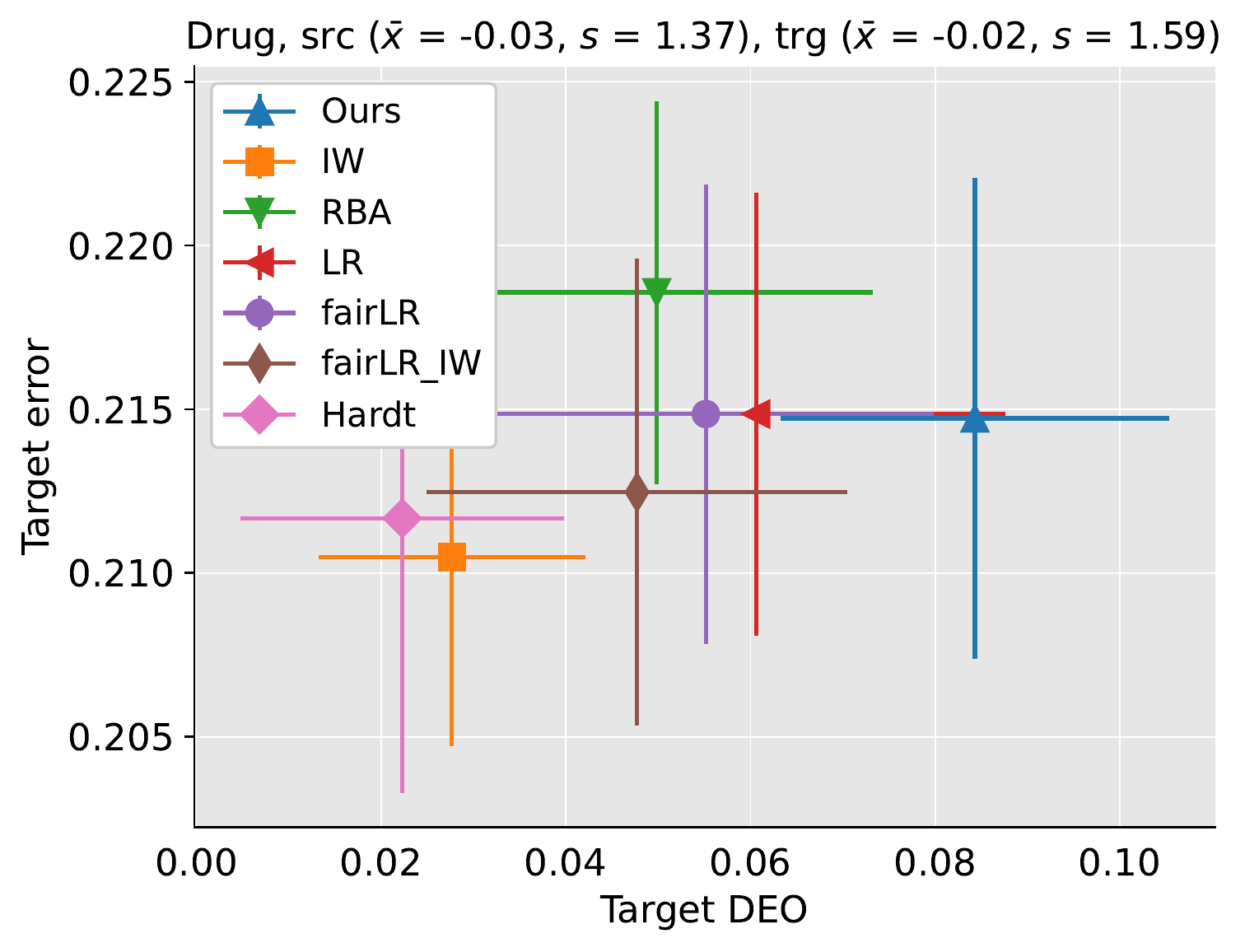} &
    \includegraphics[width=.28\textwidth,trim=20 0 0 0,clip]{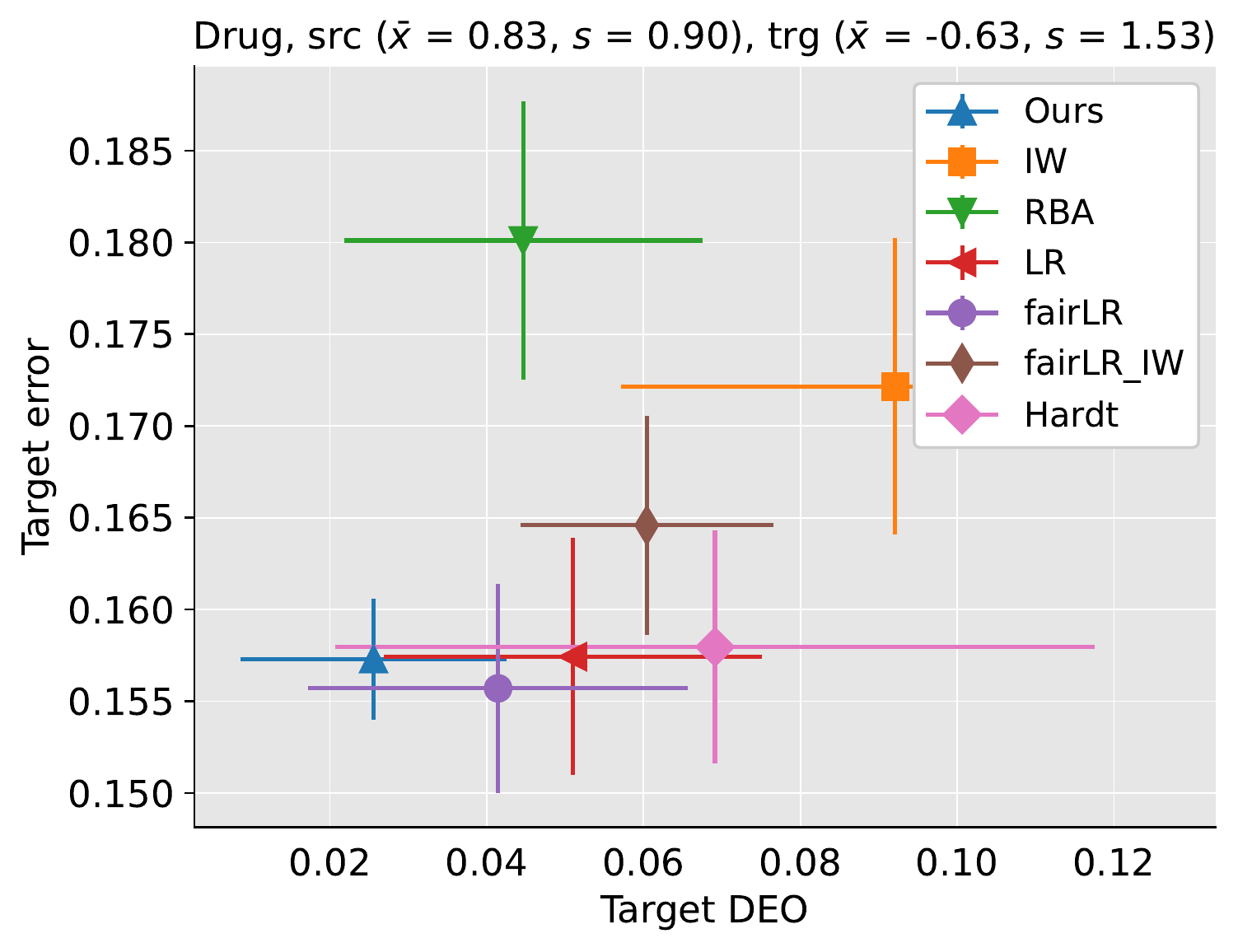} &
    \includegraphics[width=.28\textwidth,trim=20 0 0 0,clip]{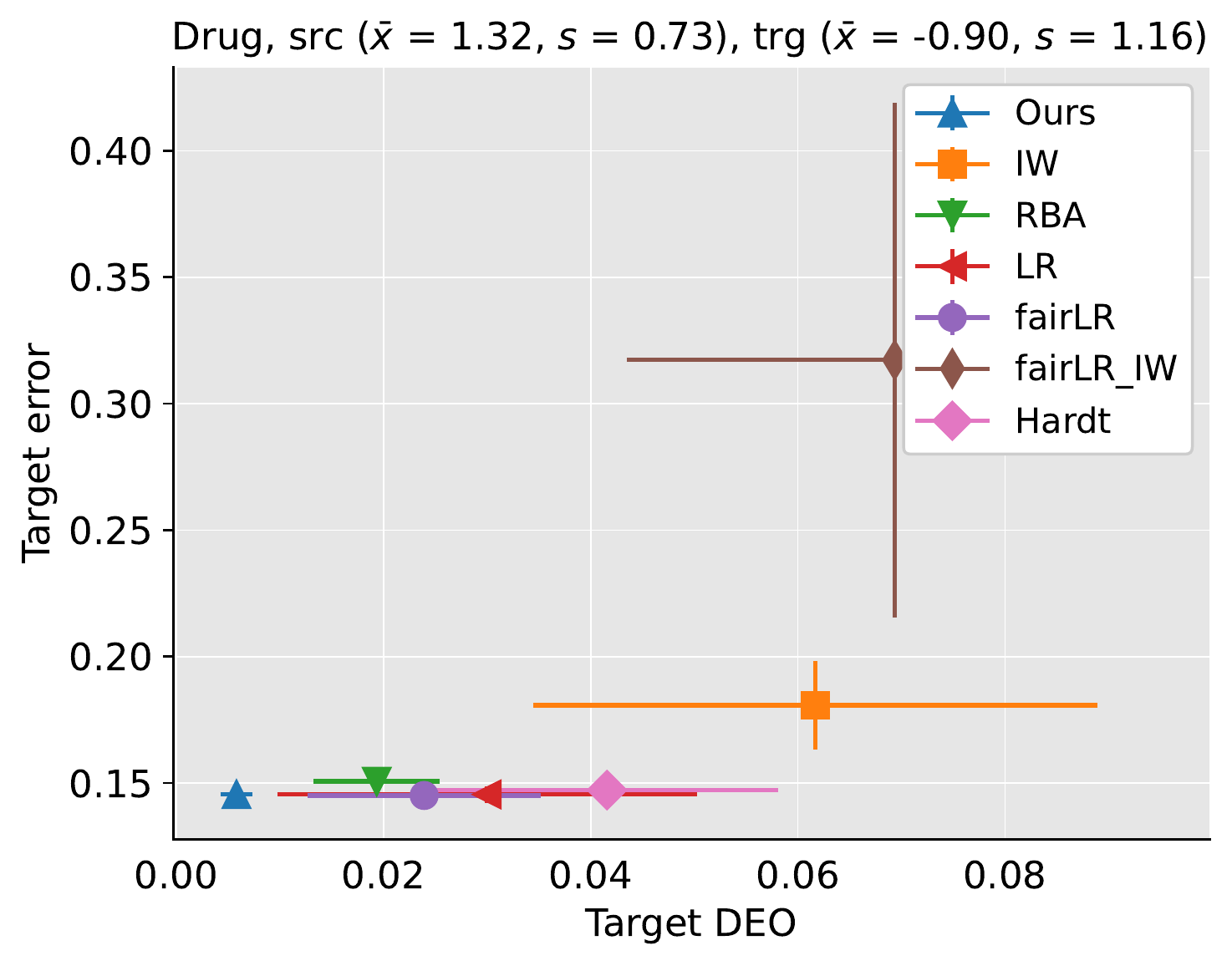}
    \\
     \includegraphics[width=.3\textwidth]{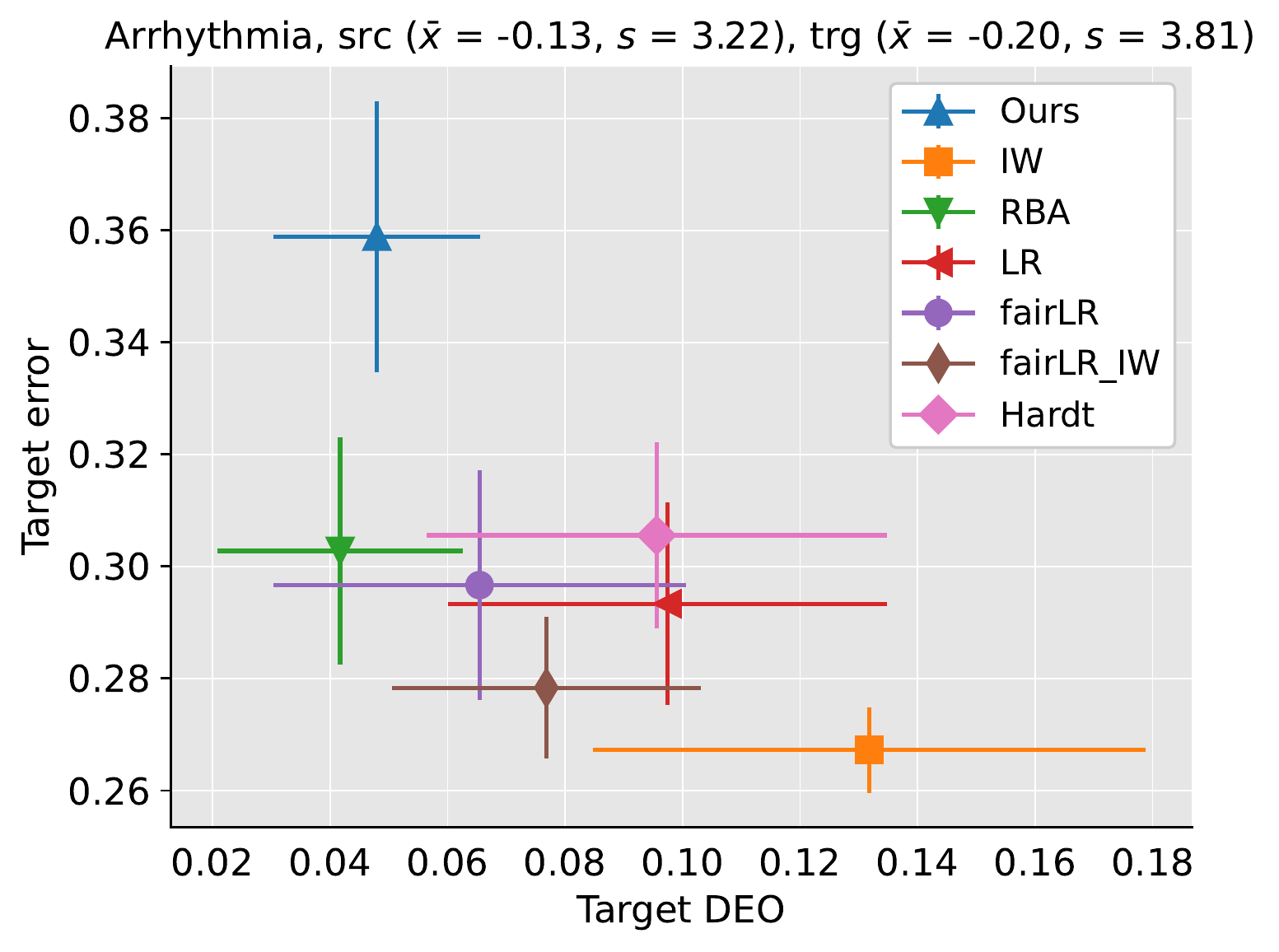} &
    \includegraphics[width=.28\textwidth,trim=20 0 0 0,clip]{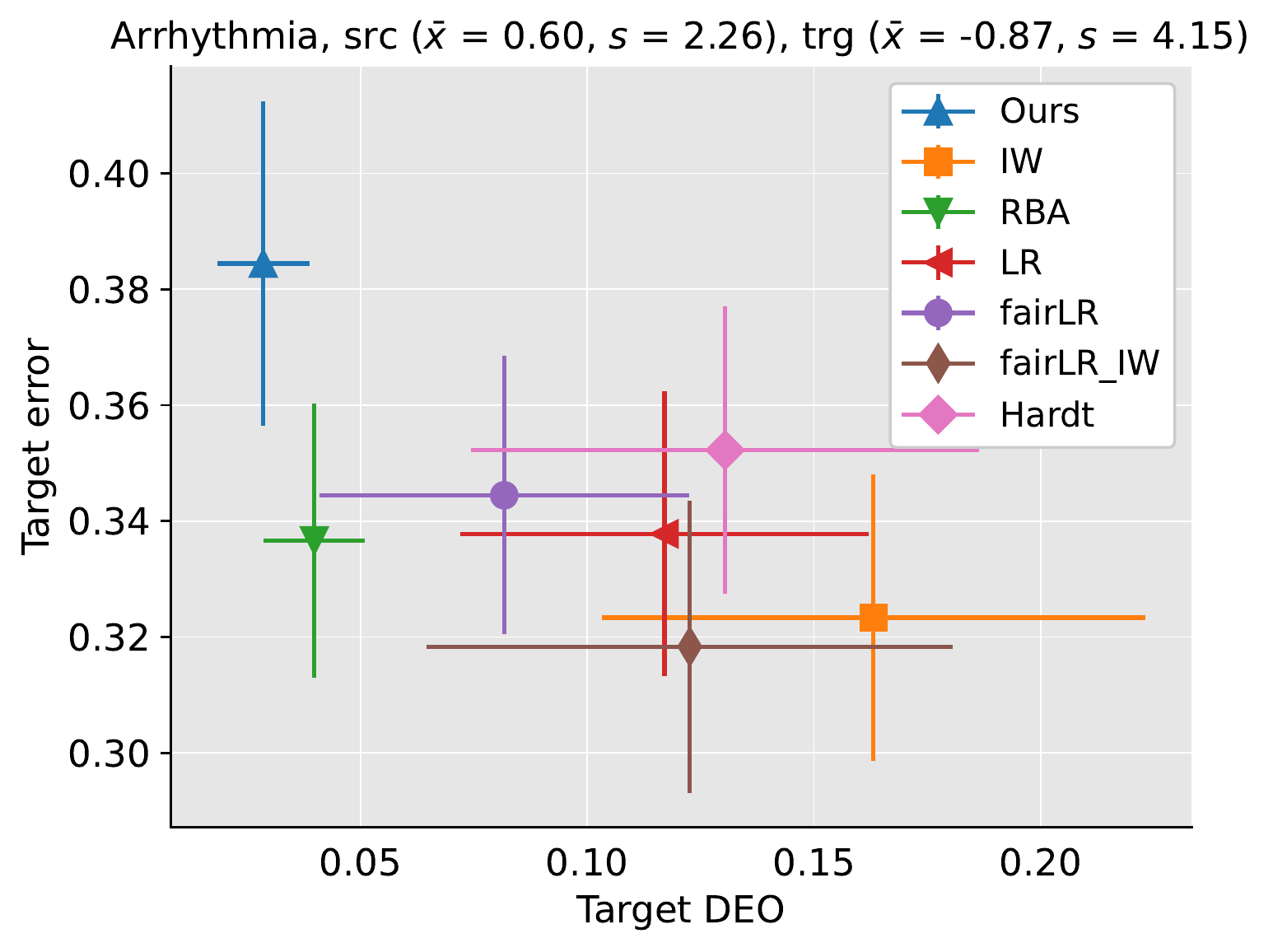} &
    \includegraphics[width=.28\textwidth,trim=20 0 0 0,clip]{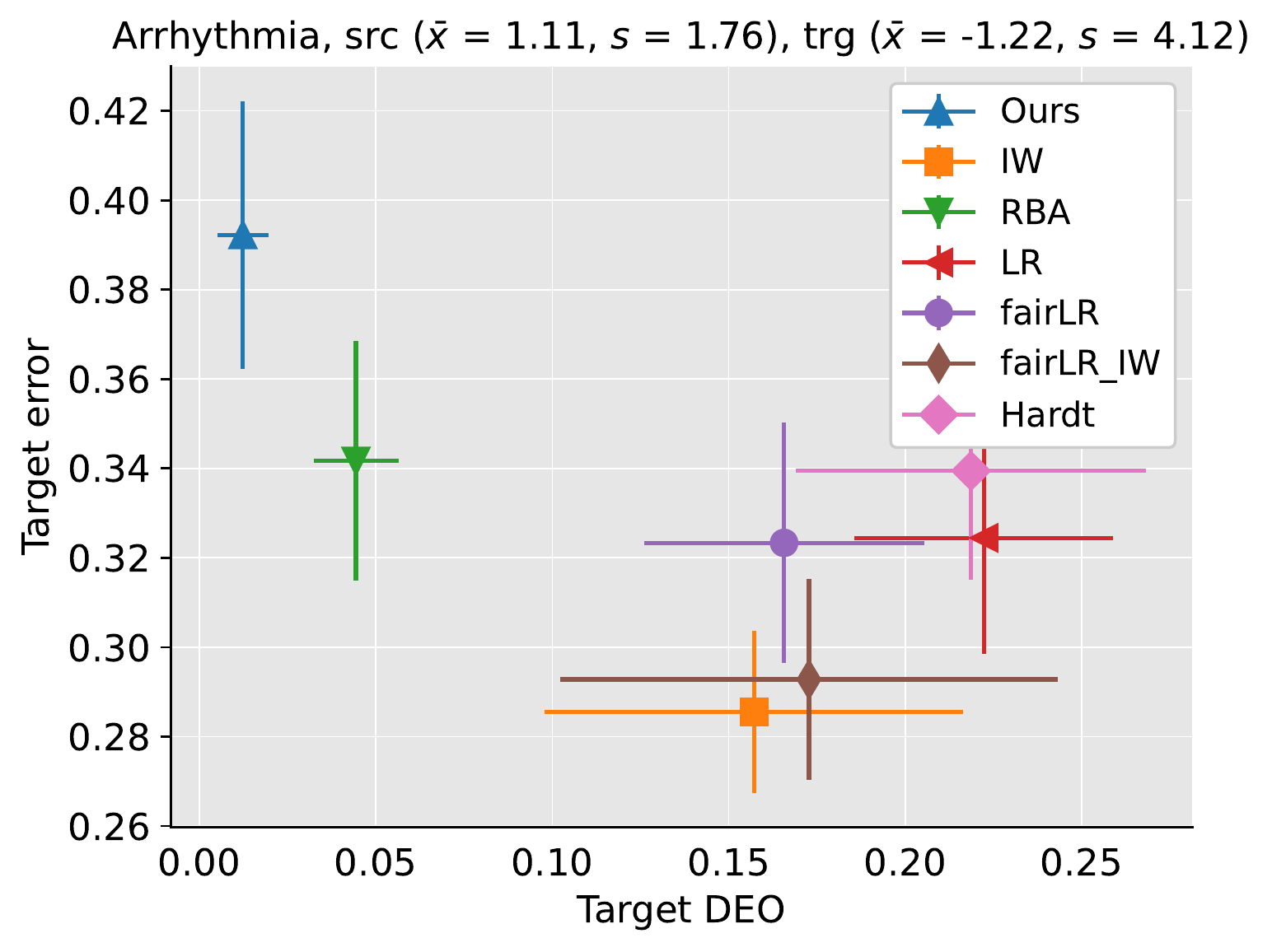}
    \\
    
    \end{tabular}
\end{center}
    \caption{Average \emph{prediction error} versus average \emph{difference of equalized opportunity} (DEO) on target samples. The bar is the 95\% confidence interval on ten random biased samplings on the first principal component of the covariates ($P_\src(x, a) \neq P_\trg(x, a)$).}
    \label{fig:results_pca}
\end{figure*}

\paragraph{Baseline methods}
We evaluate the performance of our model in terms of the trade-off between 
prediction error and fairness violation on target under various intensities of covariate shift. We focus on 
equalized opportunity as our fairness definition. We compare against the following baselines: 
\begin{itemize}
    \item {\bf Logistic Regression} ({\sc LR}) is the standard logistic regression predictor trained on source data, ignoring both covariate shift and desired fairness properties.
    \item {\bf Robust Bias-Aware Log Loss Classifier} (RBA) \cite{liu2014robust} in \eqref{eq:robustLogLossCovariateShift} 
    which accounts for the covariate shift but ignores fairness. 
    \item {\bf Sample Re-weighted Logistic Regression} ({\sc LR\_IW}) \cite{shimodaira2000improving} minimizes the re-weighted log loss on the source data, according to the importance weighting scheme \eqref{eq:importance-weighting}: 
    it only accounts for the covariate shift.
    \item {\bf Post Processing}\footnote{We use the implementation from {\tt \url{https://fairlearn.github.io}}.} ({\sc Hardt}) 
    transforms the logistic regression target output to adjust for true positive rate parity \cite{hardt2016equality}; ignores covariate shift.
    \item {\bf Fair Logistic Regression} ({\sc fairLR}) 
    also optimizes worst-case log loss subject to fairness as linear constraints with observed labels on source data \cite{rezaei2020fairness}. It accounts for fairness, but ignores the covariate shift. 
    \item {\bf Sample Re-weighted Fair Logistic Regression} ({\sc fairLR\_IW}) the fairLR method augmented with importance weighting scheme \eqref{eq:importance-weighting} in training. This baseline account for both fairness and covariate shift. 
\end{itemize}

\subsubsection{Setup}

We repeat our sampling procedure for each dataset ten times and report the average prediction error and the average difference of equalized opportunity (DEO): $|\Pbb(\Yhat = 1 | A = 1, Y = 1) - \Pbb(\Yhat = 1| A = 0, Y = 1)|$ of our predictor on the target dataset.

Unfortunately since the target distribution is assumed to be unavailable for this problem, properly obtaining optimal regularization via cross validation is not possible. We select the L2 regularization parameter by choosing the best $C$ from $\{10^{-5},10^{-4},10^{-3},10^{-2},10^{-1},1,10\}$ under the IID assumption. We use first-order features for our implementation, i.e., $\phi(\mathbf{x},y) = [x_1y, x_2y, \dots x_my]^\top$, where $m$ is the size of features.
Under the mild assumption that the estimated fairness violation by $\Qbb$ remains monotone given sufficiently expressive feature constraints on source and group marginal constraints of target, 
we find the exact zero point of the approximated violation efficiently by binary search for $\mu \in [-1.5,1.5]$ in our experiments. 

We create samplings based on three parameter settings for $(\alpha,\beta) \in \{(0,1), (1,2), (1.5,3)\}$ which is the left, middle and right column in Figure~\ref{fig:results_pca} respectively. The left column is the closest sampling to IID, as the prior sampling densities are identical. However, due to the large sampling size without replacement, the actual samples can vary quite significantly from the Gaussian prior density, as can be seen by the actual sample mean and standard deviation reported on top of each plot.   
The accuracy of estimated density ratios has a crucial impact on the performance of our model and covariate shift correction in general. We employ 2D Gaussian kernel density estimation (KDE\footnote{We use \url{https://pypi.org/project/KDEpy} package.}) with bandwidth $\sigma = 0.3$ on the first two principal components of the covariates in each sample to re-estimate the actual densities $d_i$ for the $i^\text{th}$ data point in the source and target distributions.
In order to control the magnitude of the ratios and avoid extremely large values, we normalize the densities on source and target by $p_i = (d_i + \epsilon) / (\sum_i^n (d_i + \epsilon))$, where $n$ is the size of each set. We set $\epsilon = 0.001$ for the smaller \texttt{Arrhythmia} dataset and $\epsilon = 0.01$ for the rest of the datasets. 
\subsection{Results}

Figure 
\ref{fig:results_pca} 
shows our experimental results on three samplings from close to IID (left) to mild (middle) and strong covariate shift (right). Figure \ref{fig:motivation} provides an example of these samplings on {\tt German}. 
On the {\tt COMPAS} dataset, our method consistently achieves the lowest DEO while also keeping the lowest prediction error as the shift increases, while the \textsc{Hardt} method's DEO worsen with the increasing shift. The optimal $\mu$ lies consistently close to zero on larger shifts on this dataset, which explains why \textsc{RBA} and \textsc{FairLR} are also very close to our method, indicating that the created shift was positively correlated to fairness. 
On the {\tt German} dataset, our method provides the lowest and closest to zero average DEO on all shifted samplings, with competitive prediction error compared to other baselines. As the shift intensifies DEO worsens for other baselines (except \textsc{RBA}), which shows the negative effect of covariate shift on fairness for this dataset.  
On the {\tt Drug} dataset, our method's DEO starts high. However, as the shift intensifies, our method outperforms other baselines on DEO while having equally low error as other competitive baselines. We have observed high sensitivity of results for our method and \textsc{RBA} to the accuracy of density ratios on instances of this dataset with lower shift intensity. 2D KDE provides significant improvement over 1D in this regard. We believe more accurate density estimation in higher dimensions should improve the consistency of these results.   
The samples on {\tt Arrhythmia} are much smaller and have larger standard deviation compared to the other datasets. Our method achieves the lowest fairness violation at the cost of incurring slightly higher error compared to other baselines on this dataset.  

In summary, our method achieves lowest DEO on 10 out of 12 samplings in our experiments. In four of those samplings, our method is equal with or outperforms the best baselines on prediction error as well, while being competitive for prediction error on the rest. These experiments show the overall effectiveness of our fair prediction method under general shift in the covariates.  

\section{Conclusions}

In this paper, we developed a novel adversarial approach for seeking fair decision making under covariate shift.
In contrast with importance weighting methods, our approach is designed to operate appropriately even when portions of the shift between source and target distributions are extreme.
The key technical challenge we address is the lack of labeled target data points, making target fairness assessment challenging.
We instead propose to measure approximated fairness against an worst-case adversary that is constrained by source data properties and group marginals from target. 
We incorporate fairness as a weighted penalty and tune the weighted penalty to provide fairness against the adversary.
More extensive evaluation on naturally-biased datasets and generalization of this approach to decision problems beyond binary classification are both important future directions. 

\section*{Broader Impact}

Fairness considerations are increasingly important for machine learning systems applied to key social applications. However, the standard assumptions of statistical machine learning, such as iid training and testing data, are often violated in practice. This work offers an approach for robustly seeking fair decisions in such settings and could be of general benefit to individuals impacted by alternative systems that are either oblivious or brittle to these broken assumptions. However, this work also makes a covariate shift assumption instead of accounting for more specific causal relations that may generate the shift. Practitioners should be aware of the specific assumptions made by this paper.

\section*{Acknowledgements}
This work was supported by the National Science Foundation Program on Fairness in AI in collaboration with Amazon under award No. 1939743.

\bibliography{bib}
\onecolumn

\end{document}